# Leveraging Side Observations in Stochastic Bandits

**Stéphane Caron**
Technicolor
735 Emerson St
Palo Alto, CA 94301

**Branislav Kveton**
Technicolor
735 Emerson St
Palo Alto, CA 94301

**Marc Lelarge**
INRIA–ENS
45 rue d'Ulm
75005 Paris

**Smriti Bhagat**
Technicolor
735 Emerson St
Palo Alto, CA 94301

## Abstract

This paper considers stochastic bandits *with side observations*, a model that accounts for both the exploration/exploitation dilemma and relationships between arms. In this setting, after pulling an arm $i$, the decision maker also observes the rewards for some other actions related to $i$. We will see that this model is suited to content recommendation in social networks, where users' reactions may be endorsed or not by their friends. We provide efficient algorithms based on upper confidence bounds (UCBs) to leverage this additional information and derive new bounds improving on standard regret guarantees. We also evaluate these policies in the context of movie recommendation in social networks: experiments on real datasets show substantial learning rate speedups ranging from 2.2× to 14× on dense networks.

## 1 INTRODUCTION

In the classical *stochastic multi-armed bandit* problem [4, 6], a decision maker repeatedly chooses among a finite set of $K$ actions. At each time step $t$, the action $i$ chosen yields a random reward $X_{i,t}$ drawn from a probability distribution proper to action $i$ and unknown to the decision maker. Her goal is to maximize her cumulative expected reward over the sequence of chosen actions. This problem has received well-deserved attention from the online learning community for the simple model it provides of a tradeoff between exploration (trying out all actions) and exploitation (selecting the best action so far). It has several applications, including content recommendation, Internet advertising and clinical trials.

The decision maker's performance after $n$ steps is typically measured in terms of the *regret* $R(n)$, defined as the difference between the reward of her strategy and that of an optimal strategy (one that would always choose actions with maximum expected reward). One of the most prominent algorithms in the stochastic bandit literature, UCB1 from Auer et al. [6], achieves a logarithmic (expected) regret

$$\mathbb{E}\left[R(n)\right] \leq A_{\mathsf{UCB1}} \ln n + B_{\mathsf{UCB1}}, \tag{1}$$

where $A_{\mathsf{UCB1}}$ and $B_{\mathsf{UCB1}}$ are two constants specific to the policy. This upper bound implies fast convergence to an optimal policy: the mean loss per decision after $n$ rounds is only $\mathbb{E}\left[R(n)/n\right] = O(\ln n/n)$ in expectation (which scaling is known to be optimal).

This paper considers the stochastic bandit problem *with side observations* (a setting that has been considered in [16] but for adversarial bandits, see Section 2), a generalization of the standard multi-armed bandit where playing an action $i$ at step $t$ not only results in the reward $X_{i,t}$, but also yields information on some related actions $\{X_{j,t}\}$. We also present a direct application of this scenario is advertising in social networks: a content provider may target users with promotions (*e.g.*, "20% off if you buy this movie and post it on your wall"), get a reward if the user reacts positively, but also observe her connections' feelings toward the content (*e.g.*, friends reacting by "Liking" it or not).

Our contributions are as follows. First, we consider a generalization UCB-N of UCB1 taking side observations into account. We show that its regret can be upper bounded as in (1) with a smaller $A_{\mathsf{UCB\text{-}N}} < A_{\mathsf{UCB1}}$. Then, we provide a better algorithm UCB-MaxN achieving an improved constant term $B_{\mathsf{UCB\text{-}MaxN}} < B_{\mathsf{UCB\text{-}N}}$. We show that both improvements are significant for bandits with a large number of arms and a dense reward structure, as is for example the case of advertising in social networks. We finally evaluate our policies on real social network datasets and observe substantial learning rate speedups (from 2.2× to 14×, see Section 5).

## 2 RELATED WORK

Multi-armed bandit problems became popular with the seminal paper of Robbins [19] in 1952. Thirty years later, Lai and Robbins [13] provided one of the key results in this literature when they showed that, asymptotically, the expected regret for the stochastic problem has to grow at least logarithmically in the number of steps, *i.e.*,

$$\mathbb{E}\left[R(n)\right] = \Omega(\ln n).$$

They also introduced an algorithm that follows the "optimism in the face of uncertainty" principle and decides which arm to play based on *upper confidence bounds* (UCBs). Their solution asymptotically matches the logarithmic lower bound.

More recently, Auer et al. [6] considered the case of bandits with *bounded* rewards and introduced the well-known UCB1 policy, a concise strategy achieving the optimal logarithmic bound *uniformly* over time instead of asymptotically. Further work improved the constants $A_{\mathsf{UCB1}}$ and $B_{\mathsf{UCB1}}$ in their upper bound (1) using additional statistical assumptions [3].

One of the major limitations of standard bandit algorithms appears in situations where the number of arms $K$ is large or potentially infinite; note for instance that the upper bound (1) scales linearly with $K$. One approach to overcome this difficulty is to add *structure* to the rewards distributions by embedding arms in a metric space and assuming that close arms share a similar reward process. This is for example the case in dependent bandits [18], where arms with close expected rewards are clustered.

$\mathcal{X}$-armed bandits [7] allow for an infinite number of arms $\mathbf{x}$ living in a measurable space $\mathcal{X}$. They assume that the mean reward function $\mu : \mathbf{x} \mapsto \mathbb{E}\left[X_{\mathbf{x}}\right]$ satisfies some Lipschitz assumptions and extend the bias term in UCBs accordingly. Bubeck et al. [7] provide a tree-based optimization algorithm that achieves, under proper assumptions, a regret independent of the dimension of the space.

Linear bandits [8, 20] are another example of structured bandit problems with infinitely many arms. In this setting, arms $\mathbf{x}$ live in a finite-dimensional vector space and mean rewards are modeled as linear functions of a system-wide parameter $\mathbf{Z} \in \mathbb{R}^r$, *i.e.*, $\mathbb{E}\left[X_{\mathbf{x}}\right] = \mathbf{Z} \cdot \mathbf{x}$. Near-optimal policies typically extend the notion of confidence intervals to *confidence ellipsoids*, estimated through empirical covariance matrices, and use the radius of these confidence regions as the bias term in their UCBs.

This last framework allows for contextual bandits and has been used as such in advertisement and content recommendation settings: [14] applied it to personalized news article recommendation, while [9] extended it to generalized linear models[1] and applied it to Internet advertisement. The approach in both these works is to reduce a large number of arms to a small set of numerical features, and then apply a linear bandit policy in the reduced space. Constructing good features is thus a crucial and challenging part of this process. In this paper, we do not make any assumption on the structure of the reward space. We handle the large number of arms in multi-armed bandits leveraging a phenomenon known as *side observations* which occurs in a variety of problems. This phenomenon has already been studied by Mannor et al. [16] in the case of *adversarial* bandits, *i.e.*, where the reward sequence $\{X_{i,t}\}$ is arbitrary and no statistical assumptions are made. They proposed two algorithms: ExpBan, a mix of experts and bandits algorithms based on a clique decomposition of the side observations graph, and ELP, an extension of the well-known EXP3 algorithm [5] taking the side observation structure into account. While the clique decomposition in ExpBan inspired our present work, our setting is that of *stochastic* bandits: statistical assumptions on the reward process allow us to derive $O(\ln n)$ regret bounds, while the best achievable bounds in the adversarial problem are $\widetilde{O}(\sqrt{n})$. It is indeed much harder to learn in an adversarial environment, and the methodology to address this family of problems is quite different from the techniques we use in our work.

Note that our side observations differ from *contextual side information*, another generalization of the standard bandit problems where some additional information is given to the decision maker *before* pulling an arm. Asymptotically optimal policies have been provided for this setting [22] in the case of two-armed bandits.

## 3 SIDE OBSERVATIONS

Formally, a $K$-armed bandit problem is defined by $K$ distributions $\mathcal{P}_1, \ldots, \mathcal{P}_K$, one for each "arm" of the bandit, with respective means $\mu_1, \ldots, \mu_K$. When the decision maker pulls arm $i$ at time $t$, she receives a reward $X_{i,t} \sim \mathcal{P}_i$. All rewards $\{X_{i,t}, i \in [\![1, K]\!], t \geq 1\}$ are assumed to be independent. We will also assume that all $\{\mathcal{P}_i\}$ have support in $[0, 1]$. The mean estimate for $\mathbb{E}\left[X_{i,\cdot}\right]$ after $m$ observations is $\overline{X}_{i,m} := \frac{1}{m}\sum_{s=1}^{m} X_{i,s}$. The (cumulative) regret after $n$ steps is defined by

$$R(n) := \sum_{t=1}^{n} X_{i^*,t} - \sum_{t=1}^{n} X_{I_t,t},$$

[1] in which $\mathbb{E}\left[X_{\mathbf{x}}\right] = f(\mathbf{Z} \cdot \mathbf{x})$ for some regular function $f$

| | |
|---|---|
| $K$ | # of arms |
| $X_{i,t}$ | reward of arm $i$ at time $t$ |
| $\mu_i$ | mean reward of arm $i$ |
| $\Delta_i$ | expected loss for playing arm $i$ |
| $i^*$ | index of an optimal arm |
| $\mu^*$ | mean reward of arm $i^*$ |
| $I_t$ | index of the arm played at time $t$ |
| $T_i(n)$ | # pulls to arm $i$ after $n$ steps |
| $N(i)$ | neighborhood of arm $i$ (includes $i$) |
| $O_i(n)$ | # observations for arm $i$ after $n$ steps |
| $O^*(n)$ | same for arm $i^*$ |

Table 1: Notations Summary

where $i^* = \arg\max\{\mu_i\}$ and $I_t$ is the index of the arm played at time $t$. The gambler's goal is to minimize the *expected regret* of the policy, which one can rewrite as

$$\mathbb{E}[R(n)] = \sum_{i=1}^{K} \Delta_i \mathbb{E}[T_i(n)]$$

where $T_i(n) := \sum_{t=1}^{n} \mathbf{1}\{I_t = i\}$ denotes the number of times arm $i$ has been pulled up to time $n$, and $\Delta_i := \mu^* - \mu_i$ is the expected loss incurred by playing arm $i$ instead of an optimal arm.

In the standard multi-armed bandit problem, the only information available at time $t$ is the sequence $(X_{I_s,s})_{s\leq t}$. We now present our setting with side observations. The *side observation (SO) graph* $G = (V, E)$ is an undirected graph over the set of arms $V = [\![1, K]\!]$, where an edge $i \leftrightarrow j$ means that pulling arm $i$ (resp. $j$) at time $t$ yields a side observation of $X_{j,t}$ (resp. $X_{i,t}$). Let $N(i)$ denote the *observation set* of arm $i$ consisting of $i$ and its neighbors in $G$. Contrary to previous work on UCB algorithms [13, 6], in our setting the number of observations made so far for arm $i$ at time $n$ is not $T_i(n)$ but

$$O_i(n) := \sum_{t=1}^{n} \mathbf{1}\{I_t \in N(i)\},$$

which accounts for the fact that observations come from pulling either the arm or one of its neighbors. Note that $O_i(n) \geq T_i(n)$.

A *clique* in $G$ is a subset of vertices $C \subset V$ such that all arms in $C$ are neighbors with each other. A *clique covering* $\mathcal{C}$ of $G$ is a set of cliques such that $\bigcup_{C\in\mathcal{C}} C = V$. Table 1 summarizes our notations.

### 3.1 LOWER BOUND

Before we analyze our policies, let us note that the problem we study is at least as difficult as the standard multi-armed bandit problem in the sense that, even with additional observations, the expected regret for any strategy has to grow at least logarithmically in the number of rounds. The only exception to this would be a graph where every node is neighbor with an optimal arm, a particular and easier setting that we do not study here. This observation is stated by the following Theorem:

**Theorem 1.** *Let $B^* := \arg\max\{\mu_i \mid i \in V\}$ and suppose $\bigcup_{i\in B^*} N(i) \neq V$. Then, for any uniformly good allocation rule,*[2] $\mathbb{E}[R(n)] = \Omega(\ln n)$.

*Proof.* For a set of arms $S$, we denote $N(S) := \cup_{j\in S} N(j)$. Let $i^* \in B^*$ and $v := \arg\max\{\mu_j \mid j \in V\backslash N(B^*)\}$, *i.e.*, the best arm which can not be observed by pulling an optimal arm.

First assume that $N(v) \cap N(B^*) = \emptyset$. The proof follows by comparing the initial bandit problem with side observations denoted $\mathcal{A}$ with the two-armed bandit $\mathcal{B}$ *without* side observations where the reward distributions are $\mathcal{P}^*$ for the optimal arm 1 and $\mathcal{P}_v$ for the non-optimal arm 2. To any strategy for $\mathcal{A}$, we associate the following strategy for $\mathcal{B}$: if arm $i$ is played in $\mathcal{A}$ at time $t$, play in $\mathcal{B}$: arm 1 if $i \in N(B^*)$ and get reward $X_{i^*,t}$; arm 2 if $i \in N(v)$ and get reward $X_{v,t}$; no arm otherwise.

Let $n'$ denote the number of arms pulled in $\mathcal{B}$ after $n$ steps in $\mathcal{A}$. It is clear that $n' \leq n$ and a valid strategy for $\mathcal{A}$ gives a valid strategy for $\mathcal{B}$. The expected regret incurred by arm 1 in $\mathcal{B}$ is 0, and each time arm 2 is pulled in $\mathcal{B}$, a sub-optimal arm is pulled in $\mathcal{A}$ with larger expected loss. As a consequence, $\mathbb{E}[R_{\mathcal{A}}(n)] \geq \mathbb{E}[R_{\mathcal{B}}(n')]$, where $R_{\mathcal{A}}$ (resp. $R_{\mathcal{B}}$) denotes the regret in $\mathcal{A}$ (resp. $\mathcal{B}$). By the classical result of Lai and Robbins [13], $\mathbb{E}[R_{\mathcal{B}}(n')] = \Omega(\ln n')$. Hence, if $n' = \Omega(n)$ the claim follows. If $n' = o(n)$, then sub-optimal arms are played in $\mathcal{A}$ at least $n - n'$ times so that $\mathbb{E}[R_{\mathcal{A}}(n)] = \Omega(n - n') = \Omega(n)$ and the claim follows as well.

Now assume that $N(v) \cap N(B^*) \neq \emptyset$. A valid strategy for $\mathcal{A}$ does not give a valid strategy for $\mathcal{B}$ any more, since pulling an arm in $N(v) \cap N(B^*)$ gives information on both an optimal arm and $v$, *i.e.*, both arms in $\mathcal{B}$. We need to modify slightly the two-armed bandit as follows. First, we define $u := \arg\max\{\mu_i \mid i \in N(v) \cap N(B^*)\}$ and $w := v$ if $\mu_v \geq \mu_u$ and $u$ otherwise. The reward distribution for arm 2 in $\mathcal{B}$ is now $\mathcal{P}_w$. To any strategy for $\mathcal{A}$, we associate a strategy for $\mathcal{B}$ as follows: when arm $i$ is played in $\mathcal{A}$ at time $t$, play in $\mathcal{B}$:

- $i \in N(B^*)\backslash N(v) \Rightarrow$ pull arm 1, get reward $X_{i^*,t}$;
- $i \in N(v)\backslash N(B^*) \Rightarrow$ pull arm 2, get reward $X_{w,t}$;

[2] *i.e.*, not depending on the labels of the arms, see [13]

- $i \in N(v) \cap N(B^*) \Rightarrow$ pull arms 1 and 2 in two consecutive steps, getting rewards $X_{i^*,t}$ and $X_{w,t}$;
- otherwise, do not pull any arm.

Let $n'$ denote the number of arms pulled in $\mathcal{B}$ after $n$ steps in $\mathcal{A}$. We now see that any valid strategy for $\mathcal{A}$ gives a valid strategy for $\mathcal{B}$. As in previous setting, the expected regret incurred by arm 1 in $\mathcal{B}$ is 0, and each time arm 2 is pulled in $\mathcal{B}$, a sub-optimal arm is pulled in $\mathcal{A}$ with larger expected loss. As a consequence, $\mathbb{E}[R_{\mathcal{A}}(n)] \geq \mathbb{E}[R_{\mathcal{B}}(n')]$, and we can conclude as above. $\square$

### 3.2 UPPER CONFIDENCE BOUNDS

The $\mathsf{UCB1}$ policy constructs an Upper Confidence Bound for each arm $i$ at time $t$ by adding a *bias term* $\sqrt{2\ln t / T_i(t-1)}$ to its sample mean. Hence, the UCB for arm $i$ at time $t$ is

$$\mathsf{UCB}_i(t) := \overline{X}_{i,T_i(t-1)} + \sqrt{\frac{2\ln t}{T_i(t-1)}}.$$

Auer et al. [6] have proven that the policy which picks $\arg\max_i \mathsf{UCB}_i(t)$ at every step $t$ achieves the following upper bound after $n$ steps:

$$\mathbb{E}[R(n)] \leq 8 \left( \sum_{i=1}^{K} \frac{1}{\Delta_i} \right) \ln n + \left(1 + \frac{\pi^2}{3}\right) \sum_{i=1}^{K} \Delta_i. \quad (2)$$

In the setting with side observations, we will show in Section 3.3 that a generalization of this policy yields the (improved) upper bound

$$\mathbb{E}[R(n)] \leq 8 \left( \inf_{\mathcal{C}} \sum_{C \in \mathcal{C}} \frac{\max_{i\in C} \Delta_i}{\min_{i \in C} \Delta_i^2} \right) \ln n + O(K),$$

where the $O(K)$ term is the same as in (2), and the infimum is over all possible clique coverings of the SO graph. We will detail in Section 3.3 how this bound improves on the original $\sum_i 1/\Delta_i$.

We will then introduce in Section 4 an algorithm improving on the constant $O(K)$ term (remember that the number of arms $K$ is assumed $\gg 1$). By proactively using the underlying structure of the SO graph, we will reduce it to the following finite-time upper bound:

$$\left(1 + \frac{\pi^2}{3}\right) \sum_{C \in \mathcal{C}} \Delta_C + o_{n\to\infty}(1),$$

where $\Delta_C$ is the *best* individual regret in clique $C \in \mathcal{C}$. Note that, while both constant terms were linear in $K$ in Equation (2), our improved factors are both $O(|\mathcal{C}|)$ where $|\mathcal{C}|$ is the number of cliques used to cover the SO graph. We will show that this improvement is significant for dense reward structures, as is the case with advertising in social networks (see Section 5).

### 3.3 UCB-N POLICY

In the multi-armed bandit problem with side observations, when the decision maker pulls an arm $i$ after $t$ rounds of the game, he/she gets the reward $X_{i,t}$ and observes $\{X_{j,t} \,|\, j \in N(i)\}$. We consider in this section the policy $\mathsf{UCB\text{-}N}$ where one always plays the arm with maximum UCB, and updates all mean estimates $\{\overline{X}_{j,t} \,|\, j \in N(i)\}$ in the observation set of the pulled arm $i$.

**Algorithm 1** UCB-N

```
X̄, O ← 0, 0
for t ≥ 1 do
  i ← arg max_i { X̄_i + sqrt(2 ln t / O_i) }
  pull arm i
  for k ∈ N(i) do
    O_k ← O_k + 1
    X̄_k ← X_{k,t}/O_k + (1 − 1/O_k) X̄_k
  end for
end for
```

We take the convention $\sqrt{1/0} = +\infty$ so that all arms get observed at least once. This strategy takes all the side information into account to improve the learning rate. The following Theorem quantifies this improvement as a reduction in the logarithmic factor from Equation (2).

**Theorem 2.** *The expected regret of policy UCB-N after $n$ steps is upper bounded by*

$$\mathbb{E}[R(n)] \;\leq\; \inf_{\mathcal{C}} \left\{ 8 \left( \sum_{C\in\mathcal{C}} \frac{\max_{i \in C} \Delta_i}{\Delta_C^2} \right) \ln n \right\} + \left(1 + \frac{\pi^2}{3}\right) \sum_{i=1}^{K} \Delta_i,$$

*where $\Delta_C = \min_{i\in C} \Delta_i$.*

*Proof.* Consider a clique covering $\mathcal{C}$ of $G = (V, E)$, *i.e.*, a set of subgraphs such that each $C \in \mathcal{C}$ is a clique and $V = \cup_{C \in \mathcal{C}} C$. One can define the *intra-clique regret* $R_C(n)$ for any $C \in \mathcal{C}$ by

$$R_C(n) \;:=\; \sum_{t<n} \sum_{i \in C} \Delta_i \mathbf{1}\{I_t = i\}.$$

Since the set of cliques covers the whole graph, we have $R(n) \leq \sum_{C\in\mathcal{C}} R_C(n)$. From now on, we will focus on upper bounding the intra-clique regret for a given clique $C \in \mathcal{C}$.

Let $T_C(t) := \sum_{i \in C} T_i(t)$ denote the number of times (any arm in) clique $C$ has been played up to time $t$.

Then, for any positive integer $\ell_C$,

$$R_C(n) \ \le \ \ell_C \max_{i\in C} \Delta_i + \sum_{\substack{i\in C\\ t\le n}} \Delta_i \mathbf{1}\left\{I_t = i; T_C(t-1) \ge \ell_C\right\}$$

Considering that the event $\{I_t = i\}$ implies $\{\overline{X}_{i,O_i(t-1)} + c_{t-1,O_i(t-1)} \ge \overline{X}^*_{O^*(t-1)} + c_{t-1,O^*(t-1)}\}$, we can upper bound this last summation by:

$$\sum_{\substack{i\in C\\ t<n}} \Delta_i \mathbf{1}\left\{\begin{array}{l} \overline{X}_{i,O_i(t)} + c_{t,O_i(t)} \ge \overline{X}^*_{O^*(t)} + c_{t,O^*(t)} \\ T_C(t) \ge \ell_C \end{array}\right\}$$

$$\le \sum_{\substack{i\in C\\ t<n}} \Delta_i \mathbf{1}\left\{\max_{\ell_C \le s_i \le t} \overline{X}_{i,s_i} + c_{t,s_i} \ge \min_{0\le s\le t} \overline{X}^*_s + c_{t,s}\right\}$$

$$\le \sum_{\substack{i\in C\\ t<n}} \sum_{s=0}^{t} \sum_{s_i=\ell_C}^{t} \Delta_i \mathbf{1}\left\{\overline{X}_{i,s_i} + c_{t,s_i} \ge \overline{X}^*_s + c_{t,s}\right\}$$

Now, choosing

$$\ell_C \ge \max_{i\in C} \frac{8\ln n}{\Delta_i^2} = \frac{8\ln n}{\min_{i\in C}\Delta_i^2} = \frac{8\ln n}{\Delta_C^2}$$

will ensure that $\mathbb{P}\left(\overline{X}_{i,s_i} + c_{t,s_i} \ge \overline{X}^*_s + c_{t,s}\right) \le 2t^{-4}$ for any $i \in C$ as a consequence of the Chernoff-Hoeffding bound, and following the same argument as in [6]. Hence, the overall clique regret is bounded by:

$$R_C(n) \le \ell_C \max_{i\in C} \Delta_i + \sum_{i\in C}\sum_{t=1}^{\infty} 2\Delta_i t^{-2}$$

$$\le 8\frac{\max_{i\in C}\Delta_i}{\Delta_C^2}\ln n + \left(1 + \frac{\pi^2}{3}\right)\sum_{i\in C}\Delta_i.$$

Summing over all cliques in $\mathcal{C}$ and taking the infimum over all possible coverings $\mathcal{C}$ yields the aforementioned upper bound. □

**Remark.** When $\mathcal{C}$ is the trivial covering $\{\{i\}, i \in V\}$, this upper bound reduces exactly to Equation (2). Therefore, taking side observations into account systematically improves on the baseline UCB1 policy.

# 4 UCB-MaxN POLICY

The second term in the upper bound from Theorem 2 is still linear in the number of arms and may be large when $K \gg 1$. In this section, we introduce a new policy that makes further use of the underlying reward observations to improve performances.

Consider the two extreme scenarii that can make an arm $i$ played at time $t$: it has the highest UCB, so

- either its average estimate $\overline{X}_{i,t}$ is very high, which means it is empirically the best arm to play,
- or its bias term $\sqrt{2\ln t/O_i(t-1)}$ is very high, which means one wants more information on it.

In the second case, one wants to observe a sample $X_{i,t}$ to reduce the uncertainty on arm $i$. But in the side observation setting, we don't have to pull this arm directly to get an observation: we may as well pull any of its neighbors, especially one with higher empirical rewards, and reduce the bias term all the same. Meanwhile, in the first case, arm $i$ will already be the best empirical arm in its observation set.

This reasoning motivates the following policy, called UCB-MaxN, where we first pick the arm we want to *observe* according to UCBs, and then pick in its observation set the arm we want to *pull*, this time according to its empirical mean only.

**Algorithm 2** UCB-MaxN

$\overline{\mathbf{X}}, \mathbf{n} \leftarrow \mathbf{0}, \mathbf{0}$
**for** $t \ge 1$ **do**
  $i \leftarrow \arg\max_i \left\{\overline{X}_i + \sqrt{\frac{2\ln t}{O_i}}\right\}$
  $j \leftarrow \arg\max_{j\in N(i)} \overline{X}_j$
  **pull** arm $j$
  **for** $k \in N(j)$ **do**
    $O_k \leftarrow O_k + 1$
    $\overline{X}_k \leftarrow X_{k,t}/O_k + (1 - 1/O_k)\overline{X}_k$
  **end for**
**end for**

Asymptotically, UCB-MaxN reduces the second factor in the regret upper bound (2) from $O(K)$ to $O(|\mathcal{C}|)$, where $\mathcal{C}$ is an optimal clique covering of the side observation graph $G$.

**Theorem 3.** *The expected regret of strategy UCB-MaxN after $n$ steps is upper bounded by*

$$\mathbb{E}\left[R(n)\right] \ \le \ \inf_{\mathcal{C}} \left\{ 8\left(\sum_{C\in\mathcal{C}} \frac{\max_{i\in C}\Delta_i}{\Delta_C^2}\right)\ln n + \left(1 + \frac{\pi^2}{3}\right)\sum_{C\in\mathcal{C}}\Delta_C \right\} + o_{n\to\infty}(1)$$

We will make use of the following lemma to prove this theorem:

**Lemma 1.** *Let $X_1, \dots, X_n$ and $Y_1, \dots, Y_m$ denote two sets of i.i.d. random variables of respective means $\mu$ and $\nu$ such that $\mu < \nu$. Let $\Delta := \mu - \nu$. Then,*

$$\mathbb{P}\left(\overline{X}_n > \overline{Y}_m\right) \le 2e^{-\min(n,m)\Delta^2/2}.$$

*Proof.* Note that either $\overline{X}_n < \frac{1}{2}(\mu+\nu) < \overline{Y}_m$ or one of the two events $\overline{X}_n > \frac{1}{2}(\mu+\nu)$ or $\overline{Y}_m < \frac{1}{2}(\mu+\nu)$ occurs. As a consequence, the probability $\mathbb{P}\left(\overline{X}_n > \overline{Y}_m\right)$ is lower than

$$\begin{aligned}
&\mathbb{P}\left(\overline{X}_n > \frac{\mu+\nu}{2}\right) + \mathbb{P}\left(\overline{Y}_m < \frac{\mu+\nu}{2}\right) \\
&\leq \mathbb{P}\left(\overline{X}_n - \mu > -\frac{\Delta}{2}\right) + \mathbb{P}\left(\overline{Y}_m - \nu < \frac{\Delta}{2}\right) \\
&\leq e^{-n\Delta^2/2} + e^{-m\Delta^2/2} \\
&\leq 2e^{-\min(n,m)\Delta^2/2}.
\end{aligned}$$

□

*Proof of Theorem 3.* Let $k_C := \arg\min_{i \in C} \Delta_i$ denote the best arm in clique $C$, and define $\delta_i := \Delta_i - \Delta_C$ for each arm $i \in C$. As in the beginning of our proof for Theorem 2, we can upper bound:

$$R_C(n) \leq \ell_C \max_{i \in C} \Delta_i + \sum_{\substack{i \in C \\ t < n}} \Delta_i \mathbf{1}\left\{I_t = i; T_C(t-1) \geq \ell_C\right\} \tag{3}$$

where this last summation is upper bounded by

$$\begin{aligned}
&\sum_{\substack{i \in C \\ t < n}} (\Delta_C + \delta_i) \mathbf{1}\left\{I_t = i; T_C(t-1) \geq \ell_C\right\} \\
&\leq \sum_{t<n} \Delta_C \mathbf{1}\left\{I_t = k_C; T_C(t-1) \geq \ell_C\right\} + \\
&\quad \sum_{\substack{i \in C \\ t < n}} \Delta_i \mathbf{1}\left\{I_t = i; T_C(t-1) \geq \ell_C\right\}
\end{aligned}$$

The first summation can be bounded using the Chernoff-Hoeffding inequality as before:

$$\begin{aligned}
&\sum_{t<n} \mathbf{1}\left\{I_t = k_C; T_C(t-1) \geq \ell_C\right\} \\
&\leq \sum_{t<n} \sum_{\substack{s \leq t \\ \ell_C \leq s_k \leq t}} \mathbf{1}\left\{\overline{X}_{k_C, s_k} + c_{t,s_k} > \overline{X}^*_s + c_{t,s}\right\} \\
&\leq 2 \sum_{t<n} t^{-2} \leq 1 + \frac{\pi^2}{3}
\end{aligned}$$

with an appropriate choice of $\ell_C \geq \frac{8 \ln n}{\Delta_C^2}$. As to the second summation, the fact that Algorithm 2 picks $i$ instead of $k_C$ at step $t$ implies that $\overline{X}_{i,O_i(t)} > \overline{X}_{k_C, O_{k_C}(t)}$, so

$$\begin{aligned}
R'_C(n) &:= \sum_{\substack{i \in C \\ t < n}} \Delta_i \mathbf{1}\left\{I_t = i; T_C(t-1) \geq \ell_C\right\} \\
&\leq \sum_{\substack{i \in C \\ t < n}} \Delta_i \mathbf{1}\left\{\begin{array}{l} \overline{X}_{i,O_i(t-1)} > \overline{X}_{k_C, O_{k_C}(t-1)} \\ T_C(t-1) \geq \ell_C \end{array}\right\}
\end{aligned}$$

Consider the times $\ell_C \leq \tau_1 \leq \cdots \leq \tau_{T_C(n)}$ when the clique $C$ was played (after the first $\ell_C$ steps). Then, one can rewrite $R'_C(n)$ as follows:

$$\begin{aligned}
R'_C(n) &\leq \sum_{u=\ell_C}^{T_C(n)} \sum_{i \in C} \Delta_i \mathbf{1}\left\{\overline{X}_{i,O_i(\tau_u)} > \overline{X}_{k_C, O_{k_C}(\tau_u)}\right\} \\
\mathbb{E}\left[R'_C(n)\right] &\leq \sum_{u=\ell_C}^{T_C(n)} \sum_{i \in C} \Delta_i \mathbb{P}\left(\overline{X}_{i,O_i(\tau_u)} > \overline{X}_{k_C, O_{k_C}(\tau_u)}\right)
\end{aligned}$$

After the clique $C$ has been played $u$ times, all arms in $C$ being neighbors in the side observation graph, we know that each estimate $\overline{X}_i, i \in C$ has at least $u$ samples, *i.e.*, $O_i(\tau_u) \geq u$. Therefore, using Lemma 1 with "$n = O_i(\tau_u)$" and "$m = O_{k_C}(\tau_u)$" in the previous expression yields

$$\begin{aligned}
\mathbb{E}\left[R'_C(n)\right] &\leq \sum_{i \in C} \sum_{u=\ell_C}^{T_C(n)} 2\Delta_i e^{-u\delta_i^2/2} \\
&\leq 2 \sum_{\substack{i \in C \\ \delta_i > 0}} \Delta_i \frac{1 - e^{-n\delta_i^2/2}}{1 - e^{-\delta_i^2/2}} e^{-\ell_C \delta_i^2/2},
\end{aligned}$$

where $\delta_i = \mu_i - \min_{j \in C} \mu_j$. Combining all these separate upper bounds in Equation (3) leads us to

$$\begin{aligned}
\mathbb{E}\left[R_C(n)\right] \leq\ & 8 \frac{\max_{i \in C} \Delta_i}{\Delta_C^2} \ln n + \left(1 + \frac{\pi^2}{3}\right) \Delta_C \\
&+ 2 \sum_{\substack{i \in C \\ \delta_i > 0}} \Delta_i \frac{1 - e^{-n\delta_i^2/2}}{1 - e^{-\delta_i^2/2}} \cdot \left(\frac{1}{n}\right)^{4\delta_i^2/\Delta_C^2}
\end{aligned}$$

where this last term is $o_{n \to \infty}(1)$. □

**Remark.** UCB-MaxN is asymptotically better than UCB-N: again, its upper bound expression boils down to Equation (2) when applied to the trivial covering $\mathcal{C} = \{\{i\}, i \in V\}$.

Note that our bound is achieved *uniformly over time* and not only asymptotically; we only used the $o(1)$ notation in Theorem 3 to highlight that the last term vanishes when $n \to \infty$. This term may actually be large for small values of $n$ and pathological regret distributions, *e.g.*, if some $\delta_i$ are such that $\delta_i \ll \Delta_C$. However, with distributions drawn from real datasets we observed a fast decrease: in the Flixster experiment for instance (see Section 5.3), this term was below the $(1+\pi^2/3)\Delta_C$ constant for more than 80% of the cliques after $T \sim 20K$ steps.

## 5 EXPERIMENTS

We have seen so far that our policies improve regret bounds compared to standard UCB strategies. Let us evaluate how these algorithms perform on real social network datasets. In this section, we perform three experiments. First, we evaluate the UCB-N and UCB-MaxN policies on a movie recommendation problem using a dataset from Flixster [2]. The policies are compared to three baseline solutions: two UCB variants with no side observations, and an $\varepsilon$-greedy with side observations. Second, we investigate the impact of extending side observations to friends-of-friends, a setting inspired from average user preferences on social networks that densifies the reward structure and speeds up learning. Finally, we apply the UCB-N and UCB-MaxN algorithms in a bigger social network setup with a dataset from Facebook [1].

### 5.1 DATASETS

We perform empirical evaluation of our algorithms on datasets from two social networks: Flixster and Facebook. Flixster is a social networking service in which users can rate movies. This social network was crawled by Jamali *et al.* [10], yielding a dataset with 1M users, 14M friendship relations, and 8.2M movie ratings that range from 0.5 to 5 stars. We clustered the graph using Graclus [17] and obtained a strongly connected subgraph. Furthermore, we eliminated users that rated less than 30 movies and movies rated by less than 30 users. This preprocessing step helps us to learn more stable movie-rating profiles (Section 5.2). The resulting dataset involves 5K users, 5K movies, and 1.7M ratings. The subgraph from Facebook we used was collected by Viswanath *et al.* [21] from the New Orleans region. It contains 60K users and 1.5M friendship relationships. Again, we clustered the graph using Graclus and obtained a strongly connected subgraph of 14K users and 500K edges.

### 5.2 EXPERIMENTAL SETUP

We evaluate our policies in the context of movie recommendation in social networks. The problem is set up as a repetitive game. At each turn, a new movie is sampled from a homogeneous movie database and the policy offers it at a promotional price to one user in the social network.[3] If the user rates the movie higher than 3.5 stars, we assume that he/she accepts the promotion and our reward is 1, otherwise the reward is 0. The promotion is then posted on the user's wall and we assume that all friends of that user express their opinion, *i.e.*, whether they would accept a similar offer (*e.g.*, on Facebook by "Liking" or not the promotional message). The goal is to learn a policy that gives promotions to people who are likely to accept them.

We use standard matrix factorization techniques [12] to predict users ratings from the Flixster dataset. Since the Facebook dataset does not contain movie ratings, we generated rating profiles by matching users between the Flixster and Facebook social networks. This matching is based on structural features only, such as vertex degree, the aim of this experiment being to evaluate the performances of our policies in a bigger network with similar rating distributions across vertices.

The upper bounds we derived in the analysis of UCB-N and UCB-MaxN (Theorems 2 and 3) involve the number of cliques used to cover the side observation graph; meanwhile, bigger cliques imply more observations per step, and thus a faster convergence of estimators. These observations suggest that the minimum number of cliques required to cover the graph impacts the performances of our allocation schemes, which is why we took this factor into account in our evaluation.

Unfortunately, finding a cover with the minimum number of cliques is an NP-hard problem. We addressed it suboptimally as follows. First, for each vertex $i$ in the graph, we computed a maximal clique $C_i$ involving $i$. Second, a covering using $\{C_i\}$ is found using a greedy algorithm for the SET COVER problem [11].

For each experiment, we evaluate our policies on 3 subgraphs of the social network obtained by terminating the greedy algorithm after 3%, 15%, and 100% of the graph have been covered. This choice is motivated by the following observation: the degree distribution in social networks is heavy-tailed, and the number of cliques needed to cover the whole graph tends to be on the same scale of order as the number of vertices; meanwhile, the most active regions of the network (which are of practical interest in our content recommendation scenario) are densest and thus easier to cover with cliques. Since the greedy algorithm follows a biggest-cliques-first heuristic, looking at these 3% and 15% covers allows us to focus on these densest regions.

The quality of all policies is evaluated by the *per-step regret* $r(n) := \frac{1}{n}\mathbb{E}\left[R(n)\right]$. We also computed for each plot the improvement of each policy against UCB1 after the last round $T$ (a $k\times$ improvement means that $r(T) \approx r_{\mathsf{UCB1}}(T)/k$). This number can be viewed as a speedup in the convergence to the optimal arm. Finally, all plots include a vertical line indicating the number of cliques in the cover, which is also the number of steps needed by any policy to pull every arm at least once. Before that line, all policies perform approximately the same.

[3] In accordance with the bandit framework, we further assume that the same movie is never sampled twice.

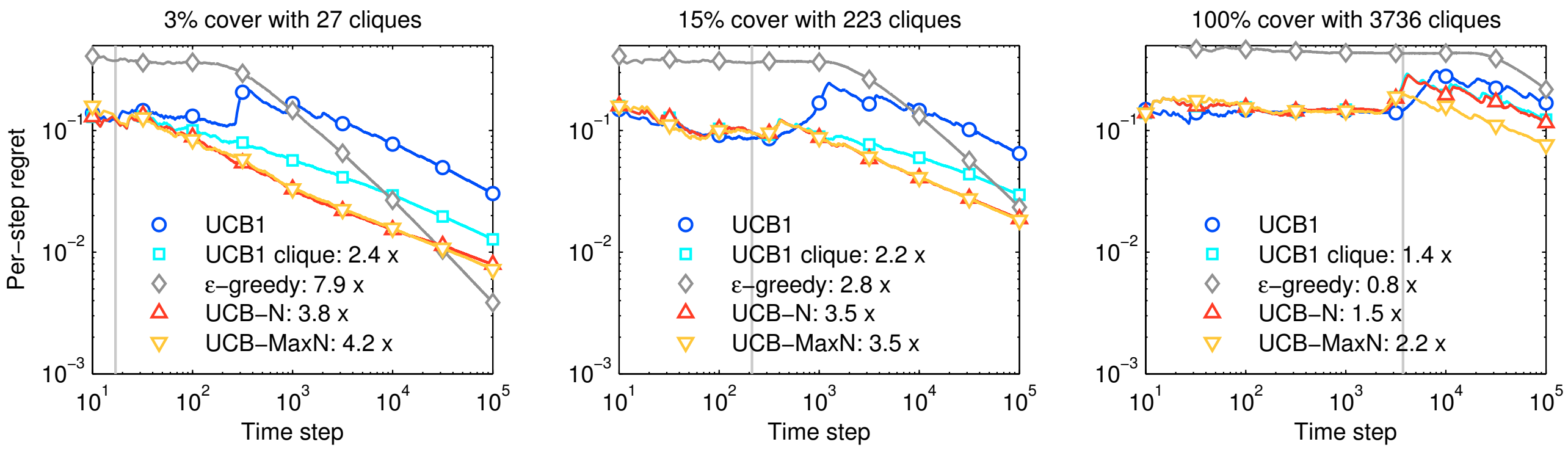

Figure 1: Per-step regret of four bandit policies on the Flixster graph.

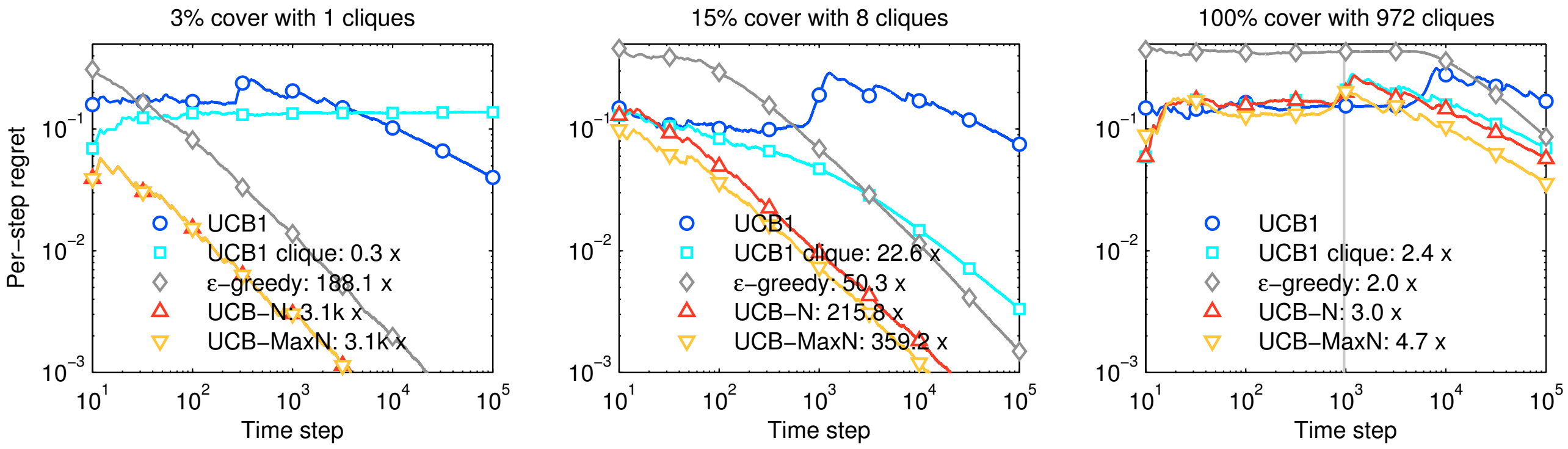

Figure 2: Per-step regret of four bandit policies on the Flixster graph with friend-of-friend side observations.

### 5.3 FLIXSTER

In this first experiment, we evaluate UCB-N and UCB-MaxN in the Flixster social network. These policies are compared to three baselines: UCB1 with no side observation, UCB1-on-cliques and $\varepsilon$-greedy. Our $\varepsilon$-greedy is the same as $\varepsilon_n$-greedy in [6] with $c = 5$, $d = 1$ and $K = |\mathcal{C}|$, which turned out to be the best empirical parametrization within our experiments. UCB1-on-cliques is similar to UCB-N, except that it updates the estimators $\{\overline{X}_k \,|\, k \in N(i)\}$ with the reward $X_{i,t}$ of the pulled arm $i$. This is a simple approach to make use of the network structure without access to side observations. As illustrated in Figure 1, we observe the following trends.

The regret of UCB-N and UCB-MaxN is significantly smaller than the regret of UCB1 and UCB1-on-cliques, which suggests these strategies successfully benefit from side observations to improve their learning rate. $\varepsilon$-greedy shows improvement as well, but its performances decrease rapidly as the size of the cover grows (*i.e.*, adding smaller cliques) compared to our strategies. Overall, the performance of all policies deteriorates with more coverage, which is consistent with the $O(K)$ and $O(|\mathcal{C}|)$ upper bounds on their regrets.

UCB-MaxN does not perform significantly better than UCB-N when the size of the cover $|\mathcal{C}|$ is small. This can be explained based on the amount of overlap between the cliques in the cover. In practice, we observed that UCB-MaxN performs better when individual arms belong to many cliques on average. For our 3%, 15%, and 100% graph cover simulations, the average number of cliques covering an arm were 1.18, 1.09, 1.76; meanwhile, the regrets of UCB-MaxN were 9%, 3%, and 33% smaller than the regrets of UCB-N, respectively.

#### 5.3.1 FRIENDS OF FRIENDS

In the second experiment we use a denser graph where side observations come from friends and friends of friends. This setting is motivated by the observation that a majority of social network users do not restrict content sharing to their friends. For instance, more than 50% of Facebook users share all their content items with friends of friends [15].

Figure 2 shows that the gap between the baselines and our policies is even wider in this new setting. This phenomenon can be explained by larger cliques; for instance, only 8 cliques are needed to cover 15% of the graph in this instance, which is 20 times less than in Section 5.3.

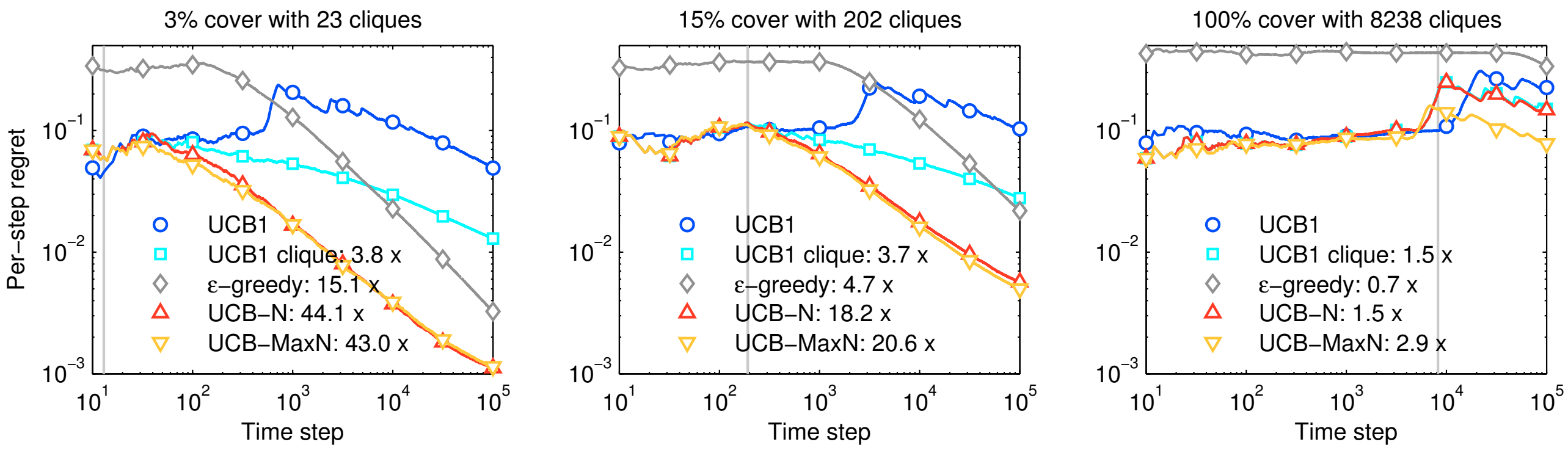

Figure 3: Per-step regret of four bandit policies on the Facebook graph.

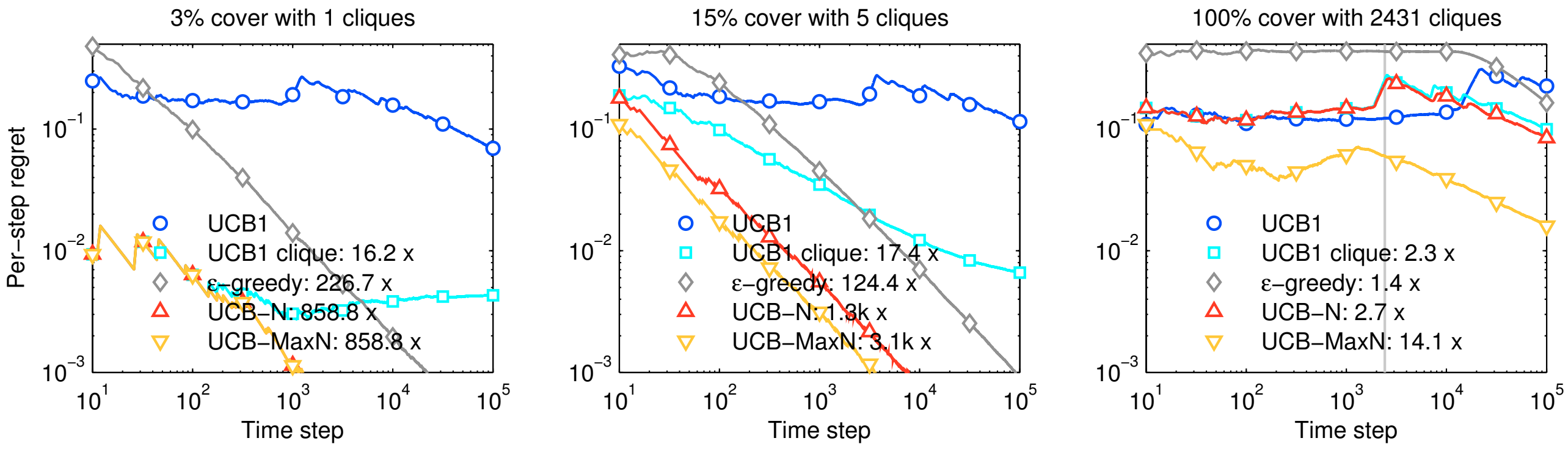

Figure 4: Per-step regret of four bandit policies on the Facebook graph with friend-of-friend side observations.

### 5.4 FACEBOOK

In the next experiment, we evaluate UCB-N and UCB-MaxN on a subset of the Facebook social network. This graph has three times as many vertices and twice as many edges as the Flixster graph. We experiment with both friends and friends-of-friends side observations.

As shown in Figures 3 and 4, compared to Figures 1 and 2, we observe much smaller regrets in this setting, essentially because the Facebook graph is denser. For instance, only 5 friend-of-friend cliques are needed to cover 15% of the graph. For this cover, the regret of UCB-MaxN is 10 times smaller than the regret of UCB1-on-cliques and UCB-N, respectively.

## 6 CONCLUSION AND FUTURE WORK

In this paper, we considered the stochastic multi-armed bandit problem with side observations. This problem generalizes the standard, independent multi-armed bandit, and has a broad set of applications including Internet advertisement and content recommendation systems. Our contribution consists in two new strategies, UCB-N and UCB-MaxN, that leverage this additional information into substantial learning rate speed-ups.

We showed that our policies reduce regret bounds from $O(K)$ to $O(|\mathcal{C}|)$, which is a significant improvement for dense reward-dependency structures. We also evaluated their performances on real datasets in the context of movie recommendation in social networks. Our experiments suggest that these strategies significantly improve the learning rate when the side observation graph is a dense social network.

So far we have focused on *cliques* as a convenient way to analyze our policies, but none of our two strategies explicitly relies on cliques (they only use the notion of neighborhood). Characterizing the most appropriate subgraph structure for this problem is still an open question that could lead to better regret bounds and inspire more efficient policies.